%% file: sbc-template.tex
\documentclass[12pt, svgnames]{article}

\usepackage{sbc-template}
\usepackage{graphicx,url}
\usepackage[utf8]{inputenc}
\usepackage{amsmath} 
\usepackage{booktabs}
\usepackage{listings}
\usepackage{soul}
\usepackage{xcolor}

\usepackage{tikz}
\usepackage{tikz-cd}
\usetikzlibrary{shapes, arrows.meta, positioning, shadows, calc}

\title{Evaluating LLMs and Prompting Strategies for Automated Hardware Diagnosis from Textual User-Reports}

\author{Carlos Caminha, Maria de Lourdes M. Silva, Iago C. Chaves, Felipe T. Brito, \\ Victor A. E. Farias, Javam C. Machado }

\address{Laboratório de Sistemas e Banco de Dados (LSBD)\\
  Departamento de Computação / Universidade Federal do Ceará \\(UFC) -- Fortaleza -- CE -- Brazil  
}

\definecolor{cyan}{HTML}{BEEBFF}
\definecolor{yellow}{HTML}{FFF59D}

\def\instnum{} 

\begin{document} 

\maketitle


\begin{abstract}  

Computer manufacturers offer platforms for users to describe device faults using textual reports such as ``My screen is flickering''. Identifying the faulty component from the report is essential for automating tests and improving user experience. However, such reports are often ambiguous and lack detail, making this task challenging. Large Language Models (LLMs) have shown promise in addressing such issues.
This study evaluates 27 open-source models (1B–72B parameters) and 2 proprietary LLMs using four prompting strategies: Zero-Shot, Few-Shot, Chain-of-Thought (CoT), and CoT+Few-Shot (CoT+FS). We conducted 98,948 inferences, processing over 51 million input tokens and generating 13 million output tokens. We achieve f1-score up to $0.76$. Results show that three models offer the best balance between size and performance: mistral-small-24b-instruct and two smaller models, llama-3.2-1b-instruct and gemma-2-2b-it, that offer competitive performance with lower VRAM usage, enabling efficient inference on end-user devices as modern laptops or smartphones with NPUs. 

\end{abstract}

\section{Introduction}
Advancements in electronic fabrication have enabled large-scale production of computer components, but faults over time remain common \cite{queiroz2016fault, queiroz2016fault2, pereira2020using}. This has increased the need for robust diagnostic systems to ensure reliability and efficiency. However, manually diagnosing these problems can be time-consuming and inefficient, making it essential for these systems to automatically identify the faulty component. By doing so, the manufacturer could also initiate an automated process to identify the suspected components, enhancing the troubleshooting process and reducing the need for manual intervention. Automating this process would enhance efficiency, lower support costs, and improve the user experience. However, this task is highly challenging because user-reported issues are typically described in free text, e.g., ``\textit{my connection keeps timing out when I try to access the internet}'', which could be ambiguous, unstructured, incomplete, and lacks technical precision \cite{maluetal2025}.

A promising solution to this problem lies in the Large Language Models (LLMs), which continue to evolve and significantly transform various domains, demonstrating their remarkable ability to perform a wide range of tasks, from natural language understanding to solving complex problems \cite{hadi2023survey}. Recent advancements have empowered these models to produce human-level responses, supporting a wide range of tasks such as programming \cite{nam2024using}, time series forecasting \cite{bastosprompt}, synthetic data generation \cite{karl2024synthetic}, classification \cite{abburi2023generative}, information extraction \cite{almeida2024evaluation}, and even complex problem-solving \cite{rasal2024llm}. 




Leveraging LLMs to analyze user-reported problem descriptions and infer potential causes presents a viable solution. These models can rapidly process ambiguous textual reports, extract relevant symptoms, and match them with likely hardware issues based on patterns learned.

For hardware manufacturers and service providers, open-source LLMs can play a key role in adopting diagnostic tools. While proprietary models such as GPT-4 and Gemini offer advanced capabilities, their reliance on external cloud services raises concerns about data privacy, competitive intelligence, and vendor dependency. Utilizing open-source models allows companies to tailor customized solutions for their proprietary diagnostic pipelines without exposing sensitive data to third parties. This approach is beneficial for businesses of all sizes, ensuring that even smaller manufacturers can integrate advanced AI models into their products without high licensing costs or data security risks.

For the effective implementation of LLM-based diagnostics, it is essential to conduct a comprehensive evaluation of open-source models across different families and parameter count. Various geopolitical and strategic factors influence model selection, as organizations may prefer specific families due to regulatory requirements, intellectual property restrictions, or regional market policies. For instance, some companies may prioritize locally developed models, such as Qwen or DeepSeek, while others may opt for Meta’s LLaMA or Google’s Gemma due to existing partnerships and technological preferences. 

In terms of parameter count, larger models tend to perform better with the trade-off of requiring more computational resources. Some companies may favor performance and deploy large models in servers with GPUs. On the other hand, given the growing integration of Neural Processing Units (NPUs) in modern devices, some organizations might prefer to deploy compact models capable of operating efficiently in edge devices. Understanding what is recommended for each case is crucial.




In this work, we tackle, by textual interaction with a computer user, the problem of predicting the computer component that is possibly causing an issue reported by the user. Our approach employs large language models using prompting engineering to map a textual user report to a possibly defective computer component. We conduct an extensive benchmark of open-source LLMs with  $27$ open-source models ranging from $1$ billion to $72$ billion parameters, along with two proprietary models. The inclusion of proprietary models helps assess the extent to which they outperform open-source models and whether the difference justifies choosing closed-source solutions for this application. Additionally, we conducted experiments using four different prompt engineering strategies: Zero-Shot, Few-Shot, Chain-of-Thought (CoT), and Chain-of-Thought combined with Few-Shot (CoT+FS). 

The contributions of this paper are listed below:

\begin{enumerate}
    \item We develop an LLM-based strategy for classifying textual user reports for possibly faulty computer components: video card, storage, network, motherboard, memory, CPU/FAN/Heatsink, battery, and audio.
    \item We conduct an extensive experimental evaluation of LLMs from various families and sizes to recommend models that offer the best balance between performance and model size.
\end{enumerate}



The remainder of this paper is structured as follows. Section~\ref{sec:related_work} reviews related work on LLM-based automated diagnosis. Section~\ref{sec:proposed_methodology} describes our methodology, detailing the dataset used, the open-source models evaluated, and the prompting strategies employed in the experiments. Section~\ref{sec:experimental_evaluation} presents the experimental setup and results, analyzing the performance of different models and prompting techniques. Finally, Section~\ref{sec:conclusion} summarizes our findings and outlines future directions for improving LLM-based hardware fault diagnosis.
\section{Related Work} \label{sec:related_work}

Despite the growing interest in LLM research, their application in hardware failure detection for automated diagnostics remains limited. Since automated diagnostics rely on accurately identifying the category of a potentially faulty component within the data, we define the scope of our related work as the use of LLMs for automated diagnosis. Standard techniques in this domain include prompt engineering~\cite{wang2024investigating, makram2024ai}, fine-tuning~\cite{zheng2024empirical}, and knowledge distillation~\cite{nathani2024knowledge}. 

\cite{li2023fault} propose a strategy for adapting LLMs, specifically Llama 2 with 7 billion parameters, to specialized fields such as fault detection and automated diagnostics in industrial machinery. Their approach involves constructing a knowledge graph tailored to the industrial domain to enhance fault diagnosis. This knowledge graph is then used to fine-tune LLMs, improving their ability to perform the target task accurately.
Similarly, \cite{tao2025llm} employs LLMs, specifically the open-source ChatGLM2 with 6 billion parameters, for bearing fault diagnostic. They convert features, such as vibration signals, into textual data, enabling the application of LLMs to this problem. In line with the previous work, the authors fine-tuned the pre-trained model using their first contribution, the textual data. On the other hand, similar to this work, \cite{maluetal2025} addresses the problem of predicting faulty hardware components using smaller models, between 22 million and 407 million parameters. The authors made an empirical comparison between different language models employing zero-, one-, and few-shot prompting strategies.


\section{Methodology} \label{sec:proposed_methodology}


This section outlines the methodology employed in our work. We consider our task a classification problem, where we aim to label a report that contains complaints about hardware issues. For instance, the model receives as input the following text ``My connection keeps timing out when I try to access the internet.'' and needs to return the ``Network'' label. We use prompt strategies in LLMs to solve this classification problem. 

\begin{figure}[!htbp]
    \centering
    \resizebox{.5\textwidth}{!}{%
        \begin{tikzpicture}[
            node distance=1cm and 1.5cm,
            block/.style={rectangle, draw, text centered, rounded corners, minimum height=2em, minimum width=7em, fill=LightSkyBlue!30},
            choice/.style={diamond, draw, aspect=2, inner sep=0pt, text centered, fill=orange!20},
            data/.style={cylinder, shape border rotate=90, aspect=0.2, draw, fill=Turquoise!20, text centered, minimum width=2cm, minimum height=1.5cm},
            arrow/.style={-{Latex[round]}, thick},
            strategy/.style={rectangle, draw, fill=Salmon!40, minimum height=2em, minimum width=7em, text centered},
            llm/.style={rectangle, draw, fill=PapayaWhip, minimum height=2em, minimum width=7em, text centered},
            group/.style={rectangle, draw, rounded corners, inner sep=0.5cm, fill=black!3},
            res/.style={draw, circle, inner sep=3pt,outer sep=0pt, fill=DarkSeaGreen!30},
            ]
        
        \node[group, double copy shadow, shadow xshift=2pt, shadow yshift=-2pt] (group) {
            \begin{tikzpicture}[node distance=1cm]
                \node[llm] (llm) {LLM};
                \node[strategy, below=of llm] (strategy) {Prompting Strategy};

                \draw[thick] (llm) -- node [circle, align=center, fill=white,inner sep=0pt,outer sep=0pt, draw=black] {+} (strategy);
            \end{tikzpicture}
        };        
        \node[block, right=of group] (evaluation) {Evaluation};
        \node[data, above=of evaluation] (dataset) {Dataset};
        \node[res, below=of evaluation] (results) {Results};
          
        
        \draw[arrow] (group) -- (group.east) -- (evaluation.west);
        \draw[arrow] (dataset) -- (dataset.south) -| (evaluation);
        \draw[arrow] (evaluation) -- (results);
        
        \end{tikzpicture}
    }
    \caption{Pipeline for evaluating LLMs in the task of faulty hardware diagnosis.}
    \label{fig:methodology-pipeline}
\end{figure}
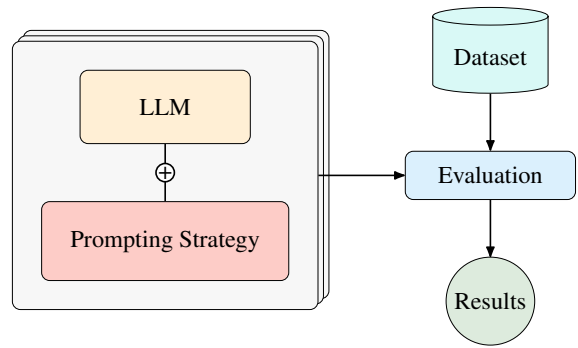

We leveraged LLMs in combination with prompting strategies to improve their performance on the classification task. To assess the effectiveness of the approach, we evaluated the model's predictions using a dataset specific to the target task, and then measured the quality of predictions using the F1-Score, a well-known metric for classification problems. Additionally, we compare various approaches, varying LLM families, LLM sizes, and prompting strategies to make an empirical evaluation and define the best LLM recommendation for this kind of task considering the balance of model's size and performance. Our methodology follows the described pipeline that is also summarized in Figure~\ref{fig:methodology-pipeline}. 


Section~\ref{sebsec:selected-llms} explains how we selected LLMs for the empirical evaluation and describes their characteristics, Section~\ref{subsec:prompt-techniques} details the prompting techniques that were combined with the LLMs, and Section~\ref{subsec:evaluation-metrics} presents the evaluation metric used to measure the predictions' quality of the approaches.

\subsection{Open Source Large Language Models}\label{sebsec:selected-llms}

To conduct this study, we selected LLMs based on their performance ranking provided by the Chatbot Arena\footnote{\url{https://chat.lmsys.org/}} up until January 1, 2025. The selection process follows two constraints: (i) Models families that provided versions with a maximum size of 72 billion parameters; (ii) Pre-quantized models in the GPT-Generated Unified Format (GGUF) with 4-bit quantization.

The first constraint ensures comparability and efficiency in test executions, while the second optimizes memory usage (VRAM) during model inference. All selected models are open-source and available on the Hugging Face platform\footnote{\url{https://huggingface.co/}}. Table~\ref{tab:model_sizes} provides an overview of the LLMs used in this study, including their source platform path, parameter size in billions, and VRAM consumption in GB. The VRAM usage is measured based on inference with a context length of eight thousand tokens.


\begin{table}[h]
\small
\centering
\caption{Overview of opensource LLMs used in this study, detailing their source platform path, parameter size (in billions), and VRAM consumption during inference with a context length of eight thousand tokens.}
\label{tab:model_sizes}
\begin{tabular}{l|cc}
\hline
\textbf{Hugging Face path} & \textbf{Size (B)} & \textbf{VRAM (GB)} \\ \hline
lmstudio-community/deepseek-r1-distill-llama-70b & 70 & 45.0 \\
lmstudio-community/deepseek-r1-distill-qwen-32b & 32 & 22.2 \\
lmstudio-community/deepseek-r1-distill-qwen-14b & 14 & 11.4 \\
lmstudio-community/deepseek-r1-distill-llama-8b & 8 & 7.8 \\
lmstudio-community/deepseek-r1-distill-qwen-1.5b & 1.5 & 3.9\\
\hline
lmstudio-community/gemma-2-27b-it & 27 & 19.2 \\
lmstudio-community/gemma-2-9b-it & 9 & 8.4 \\
lmstudio-community/gemma-2-2b-it & 2 & 4.2 \\
\hline
lmstudio-community/llama-3.3-70b-instruct & 70 & 45.0 \\
lmstudio-community/meta-llama-3-8b-instruct & 8 & 7.8 \\
lmstudio-community/llama-3.2-3b-instruct & 3 & 4.8 \\
lmstudio-community/llama-3.2-1b-instruct & 1.2 & 2.8 \\
\hline
lmstudio-community/qwen2.5-72b-instruct & 72 & 46.2 \\
bartowski/qwen2.5-32b-instruct & 32 & 22.2 \\   
lmstudio-community/qwen2.5-14b-instruct & 14 & 11.4 \\
lmstudio-community/qwen2.5-7b-instruct-1m & 7 & 7.2 \\
lmstudio-community/qwen2.5-1.5b-instruct & 1.5 & 4.1 \\
\hline
lmstudio-community/phi-4 & 14 & 11.4 \\
lmstudio-community/phi-4-mini-instruct & 3.8 & 5.3 \\
\hline
TheBloke/zephyr-7b-beta & 7 & 7.2 \\
TheBloke/stablelm-zephyr-3b & 3 & 4.8 \\
stabilityai/stablelm-zephyr-1\_6b & 1.6 & 4.0 \\
\hline
lmstudio-community/mistral-small-24b-instruct-2501 & 24 & 17.4 \\
lmstudio-community/mistral-7b-instruct-v0.3 & 7 & 7.2 \\
\hline
TheBloke/yi-34b & 34 & 23.4 \\
MaziyarPanahi/yi-9b & 9 & 8.4 \\
TheBloke/yi-6b & 6 & 6.6 \\
\hline
\end{tabular}
\end{table}

\subsection{Prompting Techniques}
\label{subsec:prompt-techniques}

We employed four distinct prompting strategies for each LLM: (i) Zero-Shot, (ii) Few-Shot, (iii) Chain-of-Thought (CoT), and (iv) CoT combined with Few-Shot. Each prompt contains specific instructions for the model to identify faulty hardware components based on user-provided descriptions, with the output being a dictionary indicating the predicted component. The prompts used to define these strategies are illustrated in Figure~\ref{fig:prompt}. The frame labeled ``Common Prompt'' is present in all prompts and outlines the target task of predicting faulty hardware components, asking the LLM to respond to user queries. The frame labeled ``Chain of Thoughts'' provides a step-by-step reasoning for the solution. The ``Few-Shot'' frame includes examples of how to respond to user queries. The prompting strategies are summarized as follows:

\begin{itemize}
    \item[i.] The Zero-Shot strategy uses only the prompts labeled as ``Common Prompt'';
    \item[ii.] Few-Shot strategy combines the ``Common Prompt'' and ``Few-shot'' frames;
    \item[iii.] CoT strategy merges the frames labeled as ``Common Prompt'' and ``Chain of Thoughts'';
    \item[iv.] The strategy that combines few-shot and CoT strategies contains all the frames described in Figure~\ref{fig:prompt}.
\end{itemize}



\lstset{
  basicstyle=\scriptsize\ttfamily, 
  breaklines=true,
  frame=single,
  rulecolor=\color{gray!50},
  backgroundcolor=\color{gray!10},
  columns=flexible,
  keepspaces=true,
}

\input{figures/prompt}

\subsection{Evaluation Metrics} \label{subsec:evaluation-metrics}

In this study, we evaluate the performance of LLMs in a classification task aimed at predicting faulty hardware components. Given the nature of our dataset and class imbalance, we adopt F1-score as the primary evaluation metric for balancing precision and recall and handle imbalanced data. This metric combines precision and recall in a single metric and it is computed as:
\(
\textit{F1 Score} = 2 \times \frac{\text{Precision} \times \text{Recall}}{\text{Precision} + \text{Recall}},
\)
\noindent where precision and recall are defined based on true positive (TP), false negative (FN), and false positive (FP) predictions, and are computed as:
\(
\textit{Precision} = \frac{TP}{TP + FP} \quad \text{and}\quad
\textit{Recall} = \frac{TP}{TP + FN}.
\)

We also evaluate the models by jointly analyzing their performance and size using the Pareto frontier, a well-established concept from multi-objective optimization~\cite{lotov2008visualizing}. The Pareto frontier captures the set of models that represent the best possible trade-offs between competing objectives. Since larger models tend to perform better than smaller ones but highly increase computational and memory costs, the competing objectives are model performance, computed by F1-score, and model size. The Pareto frontier helps to find smaller models that achieve high performance. By plotting model size on the X-axis and F1-score on the Y-axis in a 2D space, we identify the Pareto-optimal set: models for which no other model exists that is both more accurate and smaller. The Pareto frontier is thus created by connecting these non-dominated models, enabling us to visualize and compare the most efficient options with respect to the trade-off between accuracy and model complexity.

\section{Experimental Evaluation} \label{sec:experimental_evaluation}

This section details our experimental evaluation of the selected open-source large language models and prompting strategies for classifying hardware faults from textual user-reports, aiming to determine the most accurate and efficient model-strategy combinations.

\subsection{Dataset}

We evaluated our proposed combinations for classifying user-reports of hardware faults using the FACTO dataset \cite{maria2024facto}, which consists of 853 user-generated textual reports on hardware diagnosis. The dataset includes three main sources: surveys conducted with IT professionals, automated collection from specialized online forums, and synthetic generation. Each entry provides a description of a specific problem (content), the affected hardware component (label), and the source of information (source). 
    


\subsection{Evaluation Setup}

\begin{figure}[b]
    \centering
    \includegraphics[width=.8\textwidth]{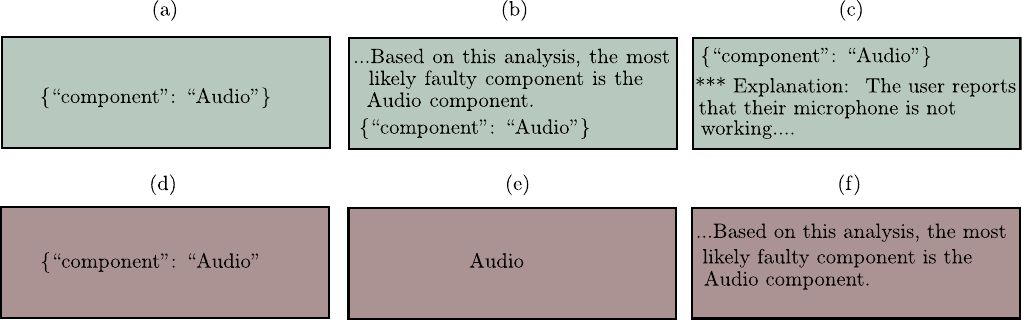}
    \caption{Examples of successful (a–c) and unsuccessful (d–f) outputs from the faulty hardware component extraction task using textual user-reports.}
    \label{fig:exemplos_saida}
\end{figure}

To evaluate the language models listed in Table~\ref{tab:model_sizes}, we employed the \texttt{llama.cpp}\footnote{\url{https://github.com/ggerganov/llama.cpp}} library. Models were limited to a context window of 8,192 tokens and used the \texttt{Q4\_K\_M} quantization strategy, a 4-bit per parameter INT4 quantization combining group quantization with mean compensation. This approach significantly reduces memory consumption without substantial performance degradation. All model layers were fully loaded onto the GPU to optimize inference speed. The experiments were performed on a local workstation equipped with a NVIDIA RTX6000 ADA GPU featuring 48 GB of VRAM.

Ensuring consistency in model outputs is critical for automated hardware fault diagnosis since structured outputs enable automated processes and integration into diagnostic pipelines. However, LLMs may generate outputs with varying levels of formatting quality, potentially complicating automated extraction tasks. To illustrate this issue, we present representative examples of outputs generated by the evaluated models in Figure \ref{fig:exemplos_saida}. It shows six examples of outputs from the component extraction task based on textual user-reports. The three examples in the first row, (a), (b), and (c), illustrate correct identification of the faulty hardware component, either as a direct JSON dictionary or embedded within correctly structured text. The second row, examples (d), (e), and (f), highlights cases where the models failed to produce the expected output. 
These cases emphasize the importance of maintaining consistent output formatting for reliable automated processing.







\subsection{Results and Discussion}

We evaluated the four prompting strategies described in Figure \ref{fig:prompt} using the LLMs detailed in Table \ref{tab:model_sizes}. In total, 98,948 inferences were executed, processing 51,641,460 input tokens and generating 13,259,092 output tokens. This extensive variation in model sizes, architectures, and prompting techniques provides a solid foundation for analyzing the impact of model size and prompting strategy on inference quality.

\begin{figure}[!h]
    \centering
    \includegraphics[width=0.9\textwidth]{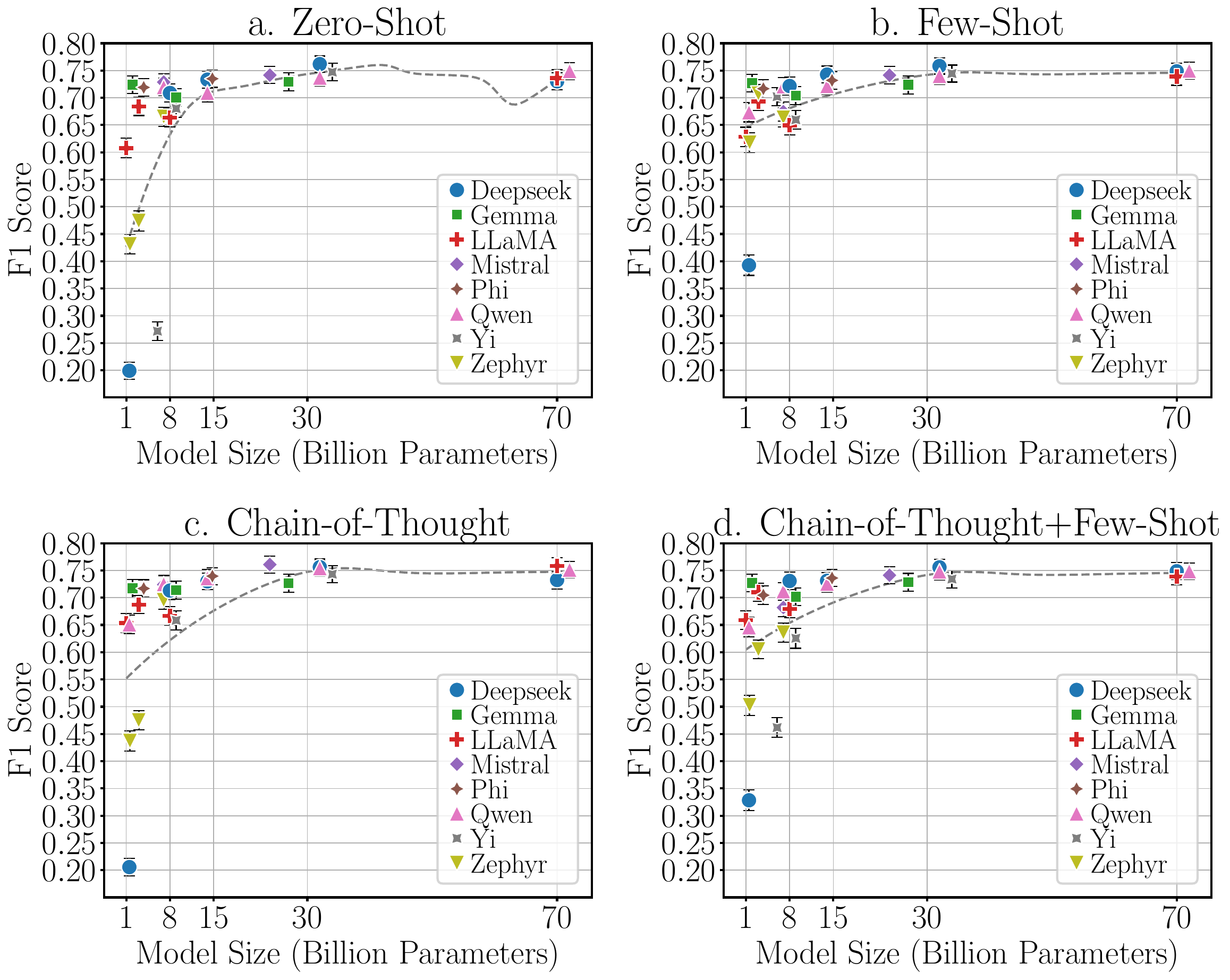}
    \caption{Comparison of F1-Scores across four prompting strategies 
    in relation to model size. The plots show the performance of various open-source LLMs, with error bars indicating the standard deviation of the F1-Score. Dashed lines represent non-parametric regressions.}
    \label{fig:painel_resultados}
\end{figure}

Figure \ref{fig:painel_resultados} presents a comparative analysis of the F1-Score results obtained for the four prompting strategies. Each plot corresponds to one of these strategies, where the horizontal axis represents model size (in billions of parameters), and the vertical axis shows the corresponding F1-Score values. Error bars indicate the standard deviation of the F1-Score, estimated using the bootstrap resampling method \cite{efron1986bootstrap}. The GPT models do not appear in the figure because their respective sizes in billions of parameters are not known.


Specifically, for each model, 1,000 random resamples with replacement were performed on the 853 inference examples from the FACTO Dataset, generating a distribution of F1-Score values. Dashed lines represent non-parametric regressions estimated using the Nadaraya-Watson method \cite{nadaraya1964estimating}. This analysis reveals a general performance improvement as model size increases. However, growth levels off near 30 billion parameters, showing limited gains beyond that point.

\begin{figure}[!h]
    \centering
    \includegraphics[width=0.75\textwidth]{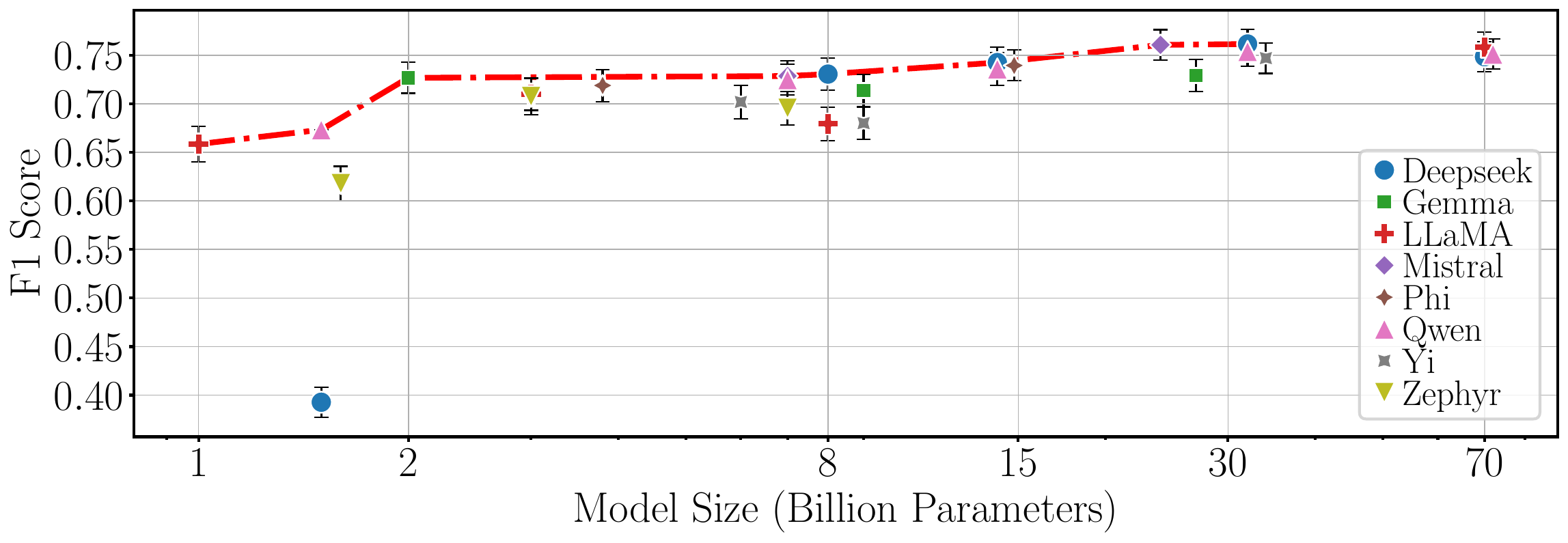}
    \caption{Pareto frontier (red dashed line) analysis of model size versus F1-score, highlighting the optimal trade-off between performance and efficiency. Models on the frontier achieve the best balance between size and performance.}
    \label{fig:pareto}
\end{figure}

We also evaluated the optimal selection of LLMs over all possible prompting strategies by analyzing the trade-off between model size and F1-score. In Figure \ref{fig:pareto}, the red dashed line represents the Pareto frontier \cite{ishizaka2013multi}, which identifies the set of models that achieve the best balance between size and performance. Models located on this frontier are considered optimal in terms of efficiency and effectiveness. Our analysis reveals that \textit{llama-3.2-1b-instruct}, \textit{gemma-2-2b-it}, and \textit{mistral-small-24b-instruct-2501} are among the most optimal models. Notably, \textit{gemma-2-2b-it} demonstrates exceptional performance, achieving an F1-score of $0.7269$ with only 2 billion parameters, compared to the F1-score of $0.7608$ achieved by \textit{mistral-small-24b-instruct-2501}, which utilizes 24 billion parameters. This highlights the efficiency of \textit{gemma-2-2b-it} in delivering competitive performance with significantly fewer parameters.

Table \ref{tab:f1_results} presents the F1-Score results, including the standard deviation, for the tested LLMs across the four prompting strategies. Bold values indicate the highest F1-Scores considering the error bars. In general, CoT often yields the best performance, particularly for larger models such as \textit{gpt-4o}, \textit{llama-3.3-70b-instruct}, and \textit{mistral-small-24b-instruct-2501}, as well as for some mid-sized models like \textit{qwen2.5-32b-instruct} and \textit{qwen2.5-14b-instruct}. In certain cases, such as \textit{deepseek-r1-distill-llama-70b}, the Few-Shot strategy (alone or combined) achieves the highest scores. 

For smaller models (especially those under $\approx 7B$ parameters), performance varies significantly. In particular, very small models like \textit{qwen2.5-1.5b-instruct} and \textit{yi-6b} exhibit extremely low F1-Scores (below 0.02) in the ZS and CoT strategies, which lack explicit examples (FS). A detailed analysis reveals that these models often fail to generate structured outputs, such as a dictionary containing the ``component'' attribute, even when correctly identifying the faulty component within unstructured text. This strongly suggests that Few-Shot prompting is essential for smaller models to ensure properly formatted outputs. In contrast, Gemma models demonstrate consistently strong performance across all strategies and sizes, indicating greater robustness to different prompting approaches.

\begin{table}[t]
\centering
\small
\caption{F1-Score results with standard deviation for the evaluated LLMs across four prompting strategies. Bold values indicate the highest F1-Scores considering the error bars for each model.}
\label{tab:f1_results}
\resizebox{.9\textwidth}{!}{%
\begin{tabular}{l|llll}
\toprule
LLM & ZS & FS & CoT & CoT+FS \\
\midrule
gpt-4o & 0.730 ± 0.016 & 0.731 ± 0.016 & \textbf{0.767 ± 0.015} & 0.746 ± 0.015 \\
gpt-4o-mini & 0.709 ± 0.017 & 0.709 ± 0.016 & \textbf{0.752 ± 0.015} & 0.729 ± 0.016 \\
\hline
deepseek-r1-distill-llama-70b & 0.730 ± 0.016 & \textbf{0.749 ± 0.016} & 0.732 ± 0.015 & \textbf{0.749 ± 0.015} \\
deepseek-r1-distill-qwen-32b & \textbf{0.762 ± 0.015} & \textbf{0.758 ± 0.015} & \textbf{0.756 ± 0.015} & \textbf{0.756 ± 0.015} \\
deepseek-r1-distill-qwen-14b & \textbf{0.733 ± 0.016} & \textbf{0.743 ± 0.015} & \textbf{0.73 ± 0.016} & \textbf{0.731 ± 0.015} \\
deepseek-r1-distill-llama-8b & 0.709 ± 0.016 & \textbf{0.722 ± 0.016} & 0.713 ± 0.016 & \textbf{0.731 ± 0.016} \\
deepseek-r1-distill-qwen-1.5b & 0.199 ± 0.016 & \textbf{0.393 ± 0.018} & 0.206 ± 0.016 & 0.328 ± 0.019 \\
\hline
gemma-2-27b-it & \textbf{0.729 ± 0.016} & \textbf{0.724 ± 0.016} & \textbf{0.726 ± 0.016} & \textbf{0.728 ± 0.016} \\
gemma-2-9b-it & \textbf{0.7 ± 0.017} & \textbf{0.704 ± 0.016} & \textbf{0.714 ± 0.016} & \textbf{0.702 ± 0.016} \\
gemma-2-2b-it & \textbf{0.724 ± 0.016} & \textbf{0.727 ± 0.016} & \textbf{0.717 ± 0.017} & \textbf{0.727 ± 0.016} \\
\hline
llama-3.3-70b-instruct & 0.736 ± 0.016 & 0.739 ± 0.015 & \textbf{0.758 ± 0.015} & 0.739 ± 0.016 \\
meta-llama-3-8b-instruct & \textbf{0.664 ± 0.018} & 0.650 ± 0.017 & \textbf{0.667 ± 0.017} & \textbf{0.679 ± 0.017} \\
llama-3.2-3b-instruct & 0.684 ± 0.016 & 0.693 ± 0.017 & 0.687 ± 0.017 & \textbf{0.710 ± 0.017} \\
llama-3.2-1b-instruct & 0.608 ± 0.018 & 0.628 ± 0.018 & \textbf{0.653 ± 0.018} & \textbf{0.659 ± 0.017} \\
\hline
qwen2.5-72b-instruct & \textbf{0.749 ± 0.015} & \textbf{0.75 ± 0.016} & \textbf{0.751 ± 0.015} & \textbf{0.749 ± 0.016} \\
qwen2.5-32b-instruct & 0.737 ± 0.015 & \textbf{0.740 ± 0.016} & \textbf{0.754 ± 0.015} & \textbf{0.748 ± 0.015} \\
qwen2.5-14b-instruct & 0.709 ± 0.016 & \textbf{0.722 ± 0.016} & \textbf{0.736 ± 0.016} & \textbf{0.726 ± 0.016} \\
qwen2.5-7b-instruct-1m & \textbf{0.72 ± 0.016} & \textbf{0.721 ± 0.016} & \textbf{0.725 ± 0.016} & \textbf{0.712 ± 0.016} \\
qwen2.5-1.5b-instruct & 0.000 ± 0.000 & \textbf{0.673 ± 0.018} & 0.651 ± 0.017 & 0.646 ± 0.018 \\
\hline
phi-4 & \textbf{0.735 ± 0.016} & \textbf{0.732 ± 0.016} & \textbf{0.74 ± 0.015} & \textbf{0.736 ± 0.016} \\
phi-4-mini-instruct & \textbf{0.719 ± 0.016} & \textbf{0.717 ± 0.017} & \textbf{0.717 ± 0.016} & \textbf{0.705 ± 0.016} \\
\hline
zephyr-7b-beta & 0.665 ± 0.017 & 0.664 ± 0.018 & \textbf{0.696 ± 0.017} & 0.636 ± 0.017 \\
stablelm-zephyr-3b & 0.474 ± 0.019 & \textbf{0.708 ± 0.017} & 0.475 ± 0.018 & 0.605 ± 0.017 \\
stablelm-2-zephyr-1\_6b & 0.431 ± 0.018 & \textbf{0.618 ± 0.018} & 0.437 ± 0.018 & 0.503 ± 0.019 \\
\hline
mistral-small-24b-instruct & 0.742 ± 0.015 & 0.741 ± 0.016 & \textbf{0.761 ± 0.015} & 0.741 ± 0.016 \\
mistral-7b-instruct-v0.3 & \textbf{0.729 ± 0.016} & 0.674 ± 0.017 & \textbf{0.724 ± 0.016} & 0.682 ± 0.017 \\
\hline
yi-34b & \textbf{0.747 ± 0.016} & \textbf{0.744 ± 0.015} & \textbf{0.743 ± 0.016} & \textbf{0.734 ± 0.016} \\
yi-9b & \textbf{0.680 ± 0.017} & 0.660 ± 0.018 & 0.658 ± 0.018 & 0.626 ± 0.018 \\
yi-6b & 0.272 ± 0.017 & \textbf{0.702 ± 0.016} & 0.013 ± 0.005 & 0.462 ± 0.019 \\
\bottomrule
\end{tabular}}
\end{table}

Other results indicate that for the task of hardware fault diagnosis based on user-generated textual descriptions, larger models such as \textit{llama-3.3-70b-instruct}, \textit{mistral-small-24b-instruct}, and \textit{qwen2.5-32b-instruct} achieve the highest performance. However, considering the feasibility of deploying compact models in real-world applications, knowledge distillation from larger models to smaller versions emerges as a promising strategy. 

\section{Conclusion} \label{sec:conclusion}

This study presented a comparative evaluation of LLMs addressed to the problem of detecting faulty hardware components, highlighting the importance of prompting strategies in optimizing failure prediction and the impact of model size on predictive performance. The Chain-of-Thought approach, particularly when combined with Few-Shot prompting, proved the most effective by enhancing structured reasoning in model responses. While the tested dataset covers only eight hardware component categories, the observed trends suggest generalizability to other failure types with similar diagnostic patterns. These findings can help manufacturers develop more efficient AI-driven diagnostic systems, reducing downtime, minimizing repair costs, and improving overall user experience. Between the evaluated models, we found that for faulty hardware prediction tasks, the \textit{mistral-small-24b-instruct} combined with the CoT strategy yielded the best results.

As a secondary contribution to this work, we can deliver potential candidates to knowledge distillation to build a specialized smaller LLM that hypothetically performs similarly to a larger model.
This distillation process transfers knowledge from larger LLMs to smaller ones, improving performance while reducing computational requirements. Among the candidates for distillation, although \textit{gemma-2-2b-it instruct} is one of the smallest models evaluated, it delivers a competitive performance compared to larger models. Additionally, larger models such as \textit{llama-3.3-70b-instruct}, \textit{mistral-small-24b-instruct}, and \textit{qwen2.5-32b-instruct} demonstrated a strong performance in the target task, making them promising candidates for knowledge distillation. Since 1B and 2B-parameter models have extremely low VRAM consumption, they enable efficient execution on modern laptops equipped with Neural Processing Units (NPUs). This allows efficient inference with minimal computational resource usage.

For future work, we propose fine-tuning compact models specifically for hardware diagnostics and developing knowledge distillation pipelines to transfer expertise from larger models. A key challenge is generating high-quality labeled examples from high-performing models to improve diagnostic generalization. As open-source LLMs and NPUs continue to advance, failure prediction diagnosis will become more accessible, enabling real-time, on-device diagnostics without relying on cloud services.

\section*{Acknowledgments}
This research was partially funded by Lenovo, as part of its R\&D investment under Brazilian Informatics Law.

\bibliographystyle{sbc}
\bibliography{sbc-template}

\end{document}

%% file: figures/prompt.tex
\begin{figure}[!h]
    \centering
    \resizebox{\textwidth}{!}{%
    \begin{tikzpicture}
        \node[draw=MistyRose!55!black, very thick, fill=MistyRose!5, rounded corners, align=justify, text width=\columnwidth, font=\tiny] (common) {
        \vspace{.2cm}\\
        \ttfamily You are an intelligent assistant capable of identifying faulty computer hardware components based on descriptions provided by their owners. Your main function is to analyze user reports of computer problems and determine which hardware component is most likely causing the issue.

        Consider the following hardware components:
        
        *   Audio: Related to sound, speakers, headphones, and microphone issues.\\
        *   Battery: Related to charging problems, battery life, or power issues if applicable.\\
        *   CPU/Fan/Heatsink: Related to processing, overheating, or fan operation problems.\\
        *   Memory: Related to random crashes, blue screen errors (BSOD), or data corruption.\\
        *   Motherboard: Related to general computer malfunctions, boot failures, or problems with ports and connectors.\\
        *   Storage: Related to hard drives (HDDs) or solid-state drives (SSDs), such as slowness, read/write errors, or corrupted files.\\
        *   Video Card: Related to graphics problems, such as visual artifacts, poor gaming performance, or issues with multiple monitors.\\
        *   Network: Related to internet connection problems, either wired (Ethernet) or wireless (Wi-Fi).\\
        
        When a user reports a problem, analyze their description and identify the most likely faulty hardware component.};
        \node[draw=MistyRose!80!gray, anchor=west, color=MistyRose!40!black, fill=MistyRose, very thick, rounded corners, anchor=east] at ([xshift=-0.5cm,yshift=-.05cm]common.north east) (common_label) {\footnotesize Common Prompt};
        \node[draw=MediumAquamarine!55!black, very thick, fill=MediumAquamarine!5, rounded corners, align=justify, font=\tiny, text width=\columnwidth, below=.3cm of common] (cot) {
        \vspace{.2cm}\\
        \ttfamily 
        First, explain your reasoning step-by-step, considering the user's description and the possible components that could be responsible.  Then, based on your reasoning, provide your final answer in the specified format. Respond ONLY with a Python dictionary where the key is "component" and the value is the name of the appropriate hardware component from the list above. Do not generate any additional text in your response under any circumstances *after the dictionary*. Ensure the component suggested is STRICTLY one from the provided list, and indicate only ONE component. The component name must MATCH EXACTLY one of the options listed above, with no variations in capitalization, spacing, or wording. Do not add or remove spaces, change the case, or alter the names in any way.
        };
        \node[draw=MediumAquamarine!55!black, anchor=west, color=MediumAquamarine!40!black, fill=MediumAquamarine!60, very thick, rounded corners, anchor=east] at ([xshift=-0.5cm,yshift=-.05cm]cot.north east) (cot_label) {\footnotesize Chain of Thoughts};
        \node[draw=Gold!55!black, very thick, fill=Gold!4, rounded corners, align=justify, font=\tiny, text width=\columnwidth, below= .3cm of cot] (fs) {
        \vspace{.2cm}\\
        \ttfamily
        Here are a few examples of how to respond to user queries:
        
        User query: My computer keeps crashing randomly, and sometimes I see a blue screen with an error message.
        Response: Let's analyze the problem. The user reports random crashes and blue screen errors (BSOD). These symptoms are most commonly associated with faulty RAM, which is listed as "Memory". Other components like the motherboard or storage *could* cause crashes, but BSODs are strongly indicative of memory issues.\\
        \{\{"component": "Memory"\}\}\\
        
        User query: I can't hear any sound coming from my speakers, and I've already checked the volume controls.
        Response: The user reports no sound from the speakers, and the volume controls are not the issue. This points directly to a problem with the audio output system.  The relevant component is "Audio".\\
        \{\{"component": "Audio"\}\}\\
        
        User query: My computer is running very slowly, and it takes a long time to open files or save documents.
        Response: The user describes slow performance, specifically when opening and saving files. This strongly suggests a problem with the storage device, where the files are read from and written to. The component is "Storage".\\ \{\{"component": "Storage"\}\}%
        };
        \node[draw=Gold!55!black, anchor=west, color=Gold!40!black, fill=Gold!30, very thick, rounded corners, anchor=east] at ([xshift=-0.5cm,yshift=-.05cm]fs.north east) (fs_label) {\footnotesize Few-shot};
        \node[draw=MistyRose!55!black, very thick, fill=MistyRose!5, rounded corners, align=justify, font=\tiny, text width=\columnwidth, below=.3cm of fs] (common2) {
        \vspace{.2cm}\\
        \ttfamily       
        Following is the user query that you detect the faulty component:\\
        User query: \{user\_query\}\\
        Response:
        };
        \node[draw=MistyRose!80!gray, anchor=west, color=MistyRose!40!black, fill=MistyRose, very thick, rounded corners, anchor=east] at ([xshift=-0.5cm,yshift=-.05cm]common2.north east) (common_label2) {\footnotesize Common Prompt};
    \end{tikzpicture}}
    \caption{Overview of prompting strategies—Zero-Shot, Few-Shot, Chain-of-Thought (CoT), and Few-Shot with CoT—to guide LLMs in identifying faulty hardware components. Each strategy integrates specific prompt frames to structure model responses based on user-provided descriptions, with outputs formatted as dictionaries predicting the faulty component.}
    \label{fig:prompt}
\end{figure}